\pdfoutput=1

\documentclass[11pt]{article}

\usepackage[final]{acl}
\usepackage{times}
\usepackage{latexsym}

\usepackage[T1]{fontenc}

\usepackage[utf8]{inputenc}

\usepackage{microtype}

\usepackage{inconsolata}

\usepackage{graphicx}

\usepackage{changepage} 
\usepackage{makecell} 
\usepackage{arydshln}
\usepackage{multirow} 
\usepackage{amsmath}
\usepackage{natbib} 
\usepackage{tabularx}
\usepackage{adjustbox}
\usepackage{subfig}
\usepackage{amssymb}
\title{GEGA: Graph Convolutional Networks and Evidence Retrieval Guided Attention for Enhanced Document-level Relation Extraction}

\author{
 \textbf{Yanxu Mao\textsuperscript{1}},
  \textbf{Xiaohui Chen\textsuperscript{2}},
 \textbf{Peipei Liu\textsuperscript{3,4 \thanks{Corresponding author.}}},
 \textbf{Tiehan Cui\textsuperscript{1}},
 \textbf{Zuhui Yue\textsuperscript{2}},
 \textbf{Zheng Li\textsuperscript{2}}
\\
 \textsuperscript{1}School of Software, Henan University, Kaifeng, China\\
 \textsuperscript{2}China Mobile Research Institute, Beijing, China\\
 \textsuperscript{3}Institute of Information Engineering, Chinese Academy of Sciences, Beijing, China\\
 \textsuperscript{4}School of Cyber Security, University of Chinese Academy of Sciences, Beijing, China\\
 \small{
   {\{maoyanxu,cuitiehan\}@henu.edu.cn} 
}
\small{
{xiaohuichen1116@gmail.com}
}\\
 \small{
   {peipliu@yeah.net} 
 }
 \small{
 {\{yuezuhui, lizheng\}@chinamobile.com}
 }
}

\begin{document}
\maketitle
\begin{abstract}
Document-level relation extraction (DocRE) aims to extract relations between entities from unstructured document text. Compared to sentence-level relation extraction, it requires more complex semantic understanding from a broader text context. Currently, some studies are utilizing logical rules within evidence sentences to enhance the performance of DocRE. However, in the data without provided evidence sentences, researchers often obtain a list of evidence sentences for the entire document through evidence retrieval (ER). Therefore, DocRE suffers from two challenges: firstly, the relevance between evidence and entity pairs is weak; secondly, there is insufficient extraction of complex cross-relations between long-distance multi-entities. To overcome these challenges, we propose GEGA, a novel model for DocRE. The model leverages graph neural networks to construct multiple weight matrices, guiding attention allocation to evidence sentences. It also employs multi-scale representation aggregation to enhance ER. Subsequently, we integrate the most efficient evidence information to implement both fully supervised and weakly supervised training processes for the model. We evaluate the GEGA model on three widely used benchmark datasets: DocRED, Re-DocRED, and Revisit-DocRED. The experimental results indicate that our model has achieved comprehensive improvements compared to the existing SOTA model.
\end{abstract}

\section{Introduction}

\begin{figure}[t]
  \includegraphics[width=\columnwidth]{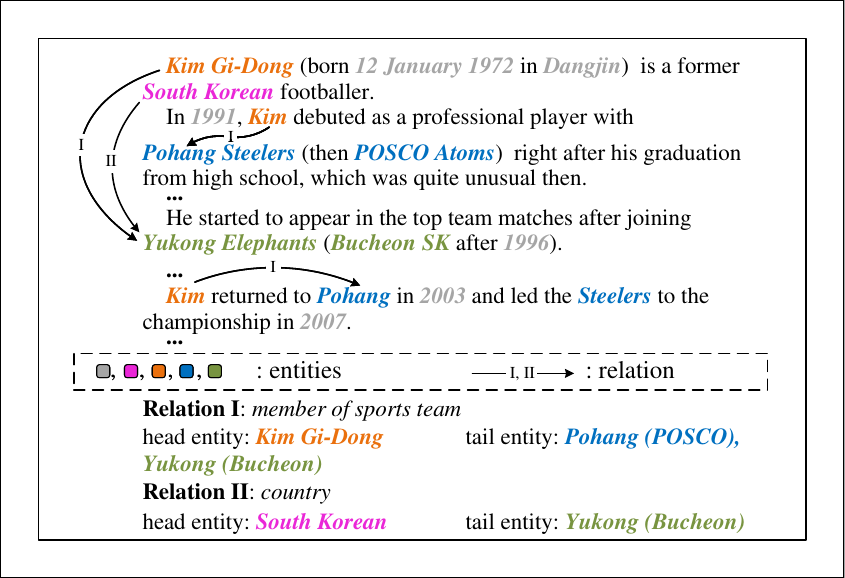}
  \caption{Examples of relations from DocRED, with entities marked in different colors, and curves indicating various relations between the entities.}
  \label{fig:docred}
  \vspace{-1ex}
\end{figure}

Relation extraction (RE) is a crucial technology used to automatically identify and classify semantic relations between entities in natural language texts. Existing relation extraction tasks can be divided into two types: sentence-level relation extraction and document-level relation extraction (DocRE) \cite{peng2017cross,verga2018simultaneously}. In sentence-level relation extraction datasets, each data entry contains only one sentence, and there is a single entity pair within the sentence for which the relation needs to be predicted. In contrast, DocRE datasets contain multiple sentences per data entry, corresponding to multiple entity pairs whose relations need to be predicted. As depicted in Figure~\ref{fig:docred}, each entity pair may appear multiple times within the document and may have different relation types \cite{yao2019docred}, necessitating the analysis of a larger contextual scope to determine the relation for each entity pair.

Furthermore, assuming there are $n$ entities in a data entry, predicting the relation for an entity pair involves pairing the anchor entity with the remaining $n$-1 entities one by one. This approach results in significant unnecessary memory overhead, as most pairs of entities do not have any relation between them.

Existing methods can be mainly divided into three categories \cite{zhou2021document} : sequence-based, graph-based, and transformer-based. Sequence-based models commonly employ pre-trained language models to produce word embeddings and character embeddings, transforming sequences of words or characters within texts into vector representations for processing \citep{ye2020coreferential, tang2020hin}. Models based on dependency graphs utilize dependency information to construct document-level graph \citep{zeng2021sire,li2021mrn,zhang2023exploringa,zhang2023exploring}, which are then processed through graph neural networks for inference. Transformer-based models utilize the self-attention mechanism to refine the representation of each word by assessing its contextual relations with all other words in the text \cite{xiao2022sais,ma2023dreeam}.

The aforementioned three types of relation extraction methods suffer from two limitations. First, relation extraction between entity pairs generally requires only a set of sentences as supporting evidence, without the need to focus on redundant irrelevant information. However, these methods utilize all the information in the long text for relation extraction. Therefore, \citet{ma2023dreeam} proposed an evidence retrieval-based relation extraction method. However, this method retrieves a list of evidence information for the entire document, resulting in poor relevance between this information and the relational entity pairs. Second, in DocRE, multiple entities are discretely distributed across different sentences or even paragraphs. Extracting their relations requires fully learning and understanding the semantics of the long text at the document-level. Existing methods rely on dependency parsing to construct multidimensional graph structures for semantic understanding and relation reasoning.However, they perform poorly when faced with complex intersecting relations due to the large number of entity pairs involved.
\\
\indent
To address the aforementioned two issues and the insufficient annotation of evidence sentences in the dataset \cite{yao2019docred}, we propose a novel model for DocRE: GEGA. This model is trained under both fully supervised and weakly supervised settings. First, we utilize a complex model (Teacher) trained with full supervision to infer over distant supervision data, extracting evidence sentences and assigning token weights. Then, we use this weight information as supervisory signals to guide the training of a simplified model (Student). Finally, we fine-tune the student model to adapt it to specific tasks and datasets, thereby improving performance. In summary, this article has two contributions:
\\
\indent
(1) We propose a novel DocRE model, GEGA\footnote{The implementation code for GEGA can be obtained from the GitHub link: : https://github.com/maoxuxu/GEGA} (\textbf{G}raph Convolutional Networks and \textbf{E}vidence Retrieval  \textbf{G}uided \textbf{A}ttention). This model combines graph structures and Transformers to retrieve evidence sentences highly relevant to the relational entity pairs from the document, guiding the attention to assign higher weights to this evidence information, thereby enhancing the performance of relation extraction.
\\
\indent
(2) Experiments conducted on the three public datasets DocRED, Re-DocRED and Revisit-DocRED show that GEGA can achieve the new SOTA\footnote{We outperform all other methods using BERT as the pretraining language model on the DocRED leaderboard. Please refer to GEGA's submission: \url{https://codalab.lisn.upsaclay.fr/competitions/365\#results}} results on document-level relation extraction compared to existing methods under the same experimental settings.

\section{Preliminary}
\subsection{Task formulation for DocRE and DocER}

Let's assume we have a document $D$ containing $n$ entities $e=\left\{e_1, e_2, \ldots, e_n\right\}$. Each entity $e_i$ in the document has a corresponding position $p_i$, and there may exist relations between entities. Our objective is to extract a set of relations $R=\left\{r_1, r_2, \ldots, r_m\right\}$ from document $D$, where each relation $r_i$ can be represented by a triple $\left(e_i, e_j, r_{i j}\right)$, with $r_{i j}$ being the relation label between entities $e_i$ and $e_j$.
Therefore, the task of DocRE can be formulated as follows:
$R=\left\{\left(e_i, e_j, r_{i j}\right) \mid e_i, e_j \in E, i \neq j\right\}$,
 $r_{i j}$ is the relation label predicted by the relation classifier based on the contextual information of entity pairs $\left(e_i, e_j\right)$.
 
 In addition, Document-level Evidence Retrieval (DocER) aims to retrieve a list of evidence sentences $evi_{[0,1,\ldots,n]}$ from document $D$ to enhance relation extraction. Nowadays, researchers have extended relation triples $\left(e_i, e_j, r_{ij}\right)$ by adding evidence sentence lists, resulting in relation quadruplets $\left(e_i, e_j, r_{ij}, evi_{[0,1,\ldots,n]}\right)$. Relations between entity pairs can be predicted solely using the sentences from the evidence lists, without relying on the entire document.

\section{Related Work}

Our work is built upon a substantial body of recent work on document-level RE and ER.\\
\noindent\textbf{DOC-Relation Extraction (RE).} Previous studies can be divided into three major categories \cite{zhou2021document}:

\indent\textsl{Sequence-based methods}. 
\citet{zeng2014relation,cai2016bidirectional,tang2020hin,yao2019docred}, and \citet{sorokin2017context} use methods such as Conditional Random Fields (CRF) or Recurrent Neural Networks (RNN, \cite{cho2014learning}) to accurately identify and label entities in text and predict the relations between these entities by learning the contextual sequential semantic information of each word. For example, \citet{tang2020hin} use different neural network architectures to perform sequence encoding of the entire document to learn the semantic representation of entity pairs and extract the relations between them. \citet{yao2019docred} employ BiLSTM to simultaneously consider the forward and backward information in the text sequence, learning and understanding the different semantic paths from one entity to another, thus inferring the relations between entity pairs.


\indent\textsl{Graph-based methods}. 
\citet{velivckovic2018graph,christopoulou2019connecting,sahu2019inter} model entities and relations in documents as graph structures, then use Graph Neural Networks (GNNs) to learn the relations between entities. This approach effectively leverages structural information and global context among entities. Additionally, researchers have proposed hard pruning and soft pruning strategies for dependency tree structures to optimize the model's speed and storage efficiency. \citet{zhang2018graph} and \citet{mandya2020graph} use hard pruning strategies to retain words near the shortest path between two entities, maximizing the removal of irrelevant content while integrating relevant information. \citet{guo2019attention} proposed a soft pruning method that directly takes the full dependency tree as input and automatically learns how to selectively focus on relevant substructures that are useful for the relation extraction task. Subsequently, \citet{li2021mrn,nan2020reasoning} proposed refined strategies to enhance cross-sentence relation reasoning by automatically inducing latent document-level graphs. This strategy allowing the model to incrementally aggregate relevant information for both local and global reasoning.

\indent\textsl{Transformer-based methods}. This approach does not use any graph structures but instead adapts to the document-level relation extraction task by fine-tuning pre-trained models \cite{wang2019fine}. \citet{ye2020coreferential} introduced a copy-based training objective into the basic pre-trained language model, enabling the model to better capture coreference information. \citet{tang2020hin} employed a hierarchical aggregation method to obtain reasoning information at different granularities at the document-level. \citet{zhou2021document} addressed the multi-label and multi-entity issues in relation extraction datasets through adaptive thresholds and local context pooling.

\noindent\textbf{DOC-Evidence Retrieval (ER).} Currently, a few studies have investigated the importance of evidence information in document-level relation extraction tasks. \citet{yao2019docred} directly incorporated evidence sentence instances supporting entity relations into a new dataset. However, in the absence of evidence sentences, evidence information needs to be generated through an Evidence Retrieval (ER) task. \citet{huang2021three} employed heuristic rules to select informative sets of paths from the entire document to discover evidence sentences and further optimized relation extraction by combining BiLSTM. \citet{ma2023dreeam} integrated evidence information into a Transformer-based DocRE system by directly guiding attention, without introducing any additional trainable parameters for the ER task. Compared to our work, they did not incorporate graph neural networks for end-to-end learning to derive the advantages of attention weights. In contrast, GEGA provides a more reliable allocation of weights for evidence information by constructing a fully connected graph and its corresponding fully connected matrix to learn structured information.
\section{Methodology}

\begin{figure*}[t]
\centering 
  \includegraphics[width=0.76\textwidth]{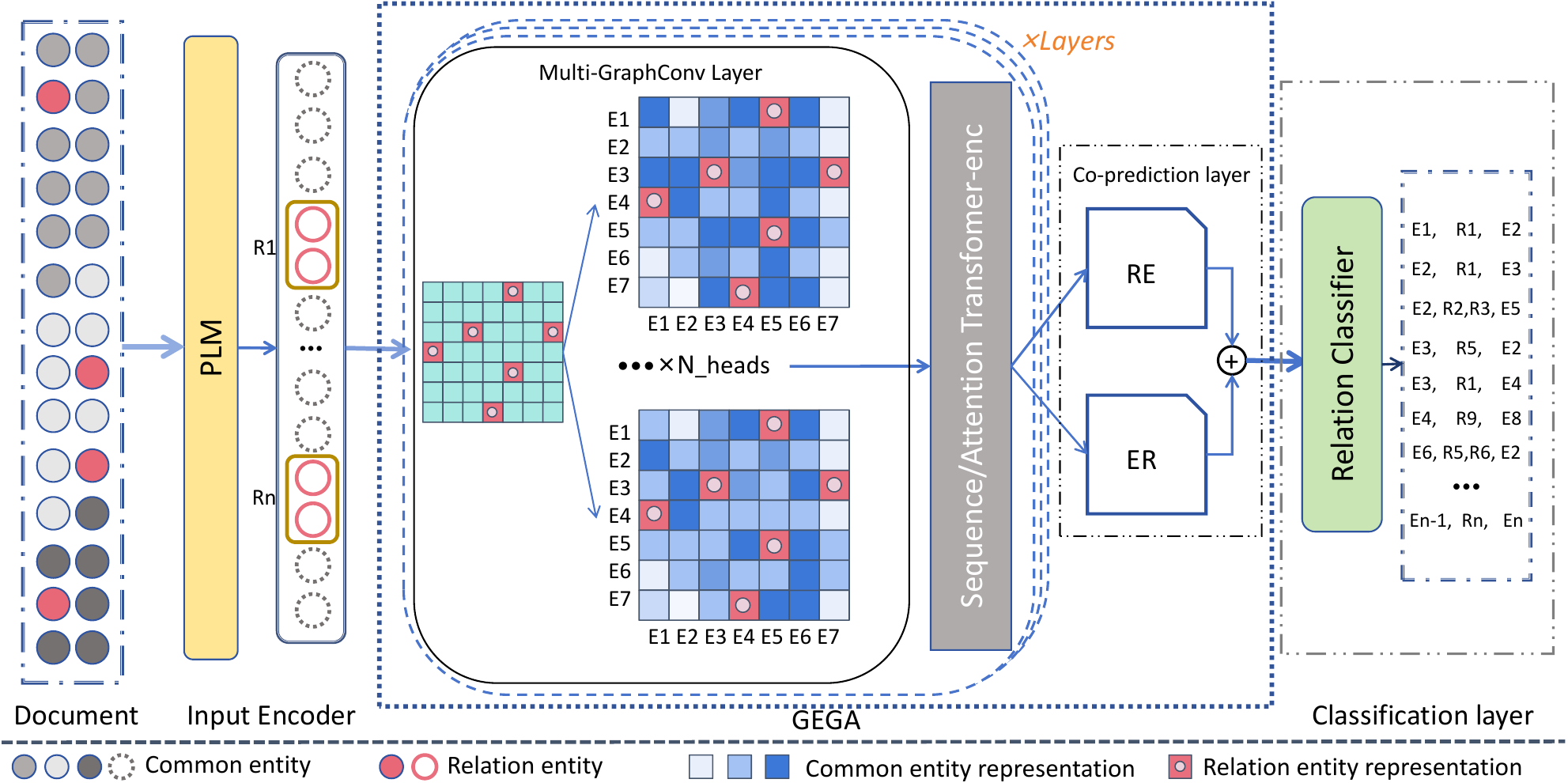}
  \caption{The overall architecture of our method. The gray circles with different depths belong to different sentences, and the color depth of the square is the basis to distinguish the attention weight score.}
  \label{fig:gega}
\end{figure*}

This section elucidates the main framework of the proposed method, illustrated in Figure~\ref{fig:gega}, the model can be segmented into tripartite: Input Encoder Layer, GEGA Module and Classification Layer.

\subsection{Input Encoder Layer}This work adheres to the methodology employed in prior research to incorporate specialized markers $[CLS]$ and $[SEP]$ at the begin and end of a designated document $doc$=$\left[sent_N\right]_{N=1}^{L}$ for the purpose of outlining the document's boundaries, where $sent$=$\left[t_n\right]_{n=1}^{l}$. Subsequently, input the document into the pre-trained BERT model. When the input length exceeds 512, the document is divided into two overlapping segments\footnote{The length of each data in the publicly available datasets is less than 1024.}. The first segment has a length of 512 and the second segment comprises the difference between the total input length and 512. Ultimately, the contextual embedding representation $H$ and attention matrix $A$ of the token are derived:

{\small
\begin{align}
\begin{aligned}
\boldsymbol{(H,A)}&=\left[\boldsymbol{({h}_1,{a}_1)}, \boldsymbol{({h}_2,{a}_2)}, \ldots, \boldsymbol{({h}_{l\times L},{a}_{l\times L})}\right] \\ &=\operatorname{BERT}([doc])
\end{aligned}
\end{align}
}where $l$ is the length of the sentence (i.e., the number of tokens), and $L$ is the length of the input document (i.e., the number of sentences).

For each entity, the efficacy of the max pooling function is pronounced when the inter-entity relations are explicitly articulated. Nevertheless, in the context of this study, the relations among entities remain ambiguous. It is understood that an entity may be referenced by one or several mentions, and a mention might uniquely identify an entity or fail to ascertain a definite corresponding entity. This necessitates the calculation of an entity's embedding based on the embeddings of each associated mention. Following the methodology of \cite{jia2019document}, a soft version of the max function $LogSumExp$ $(LSE)$ is utilized to compute the embeddings of entities:


\vspace{-2ex}
\begin{align}
\begin{aligned}
\scalebox{0.85}{$
e_{\text {emb}}=LSE\left(h_1, \ldots, h_n\right)=\log \sum_{i=1}^{\left|\mathcal{M}_e\right|} \exp \left(\boldsymbol{h}_{i \in e}\right)$}
\end{aligned}
\end{align}
where $e$ is an entity comprising multiple mentions, $\left|\mathcal{M}_e\right|$ is the number of mentions for entity $e$.

\subsection{GEGA Module}
The GEGA module comprehends four parts: the Attention Concentration Layer, the Multi-GraphConv Layer, the Transformer-enc Layer, and the Collaborative Prediction Layer.

\subsubsection{Attention Concentration Layer}

We employ Attention Concentration Layer to transform the initial dependency tree into a fully connected weighted graph based on the dependency relations within the sentence. This approach can be construed as a soft pruning strategy \cite{xu2015show} juxtaposed with the conventional hard pruning strategy \cite{guo2019attention}. By assigning weights to the sequence data, as opposed to outright deletion, a greater amount of contextual information can be preserved, thereby fostering enhancements in module efficacy. 
Subsequently, we utilize the multi-head attention mechanism, wherein the input vector is mapped to several heads using a linear transformation layer to produce an adjacency matrix with varied weight distributions, denoted as $\tilde{\boldsymbol{A}}^{(\mathbf{\operatorname{head}_i})}$=$Attention\left(Q W_i^Q, K W_i^K\right)$. We employ parallel computing to expedite the computational process.

\begin{align}
\begin{aligned}
\scalebox{0.96}{$
\tilde{\boldsymbol{A}}^{(\mathbf{\operatorname{head}_i})}=softmax\left(\frac{Q {W}_i^Q \times\left(K {W}_i^K\right)^T}{\sqrt{d}}\right)
$}
\end{aligned}
\end{align}where $Q,K \in \mathbb{R}^{N \times d_{\text{model}}}$, $N$ is the length of the sequence, $d_{\text{model}}$ is the dimensionality of the input feature, and $W_i^Q, W_i^K\in \mathbb{R}^{d_{\text{model}} \times d_k}$
 is the weight parameter associated with the linear transformation.


\subsubsection{Multi-GraphConv Layer}
\begin{figure}[t]
\centering 
  \includegraphics[width=\columnwidth]{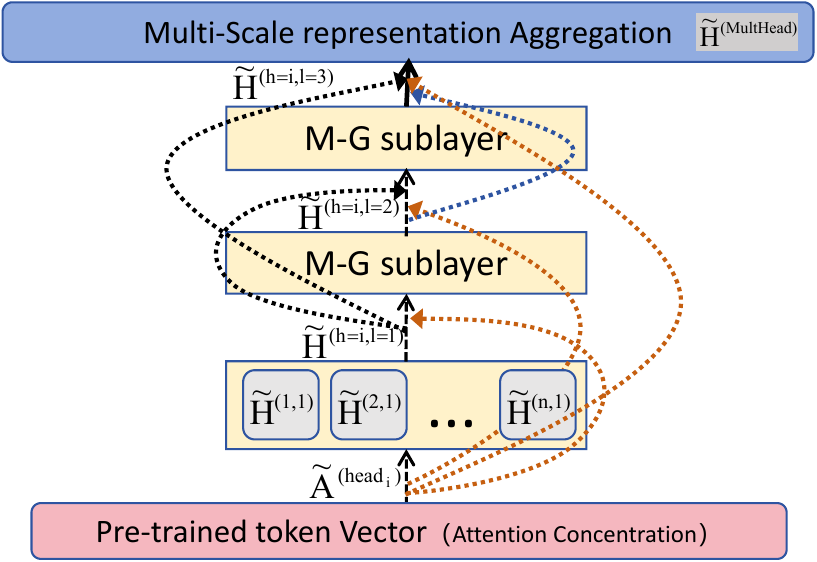}
  \caption{The overall architecture diagram of Multi-GraphConv (M-G) Layer includes three sub layers, each containing $n$ heads.}
  \label{fig:fig3}
    \vspace{-1ex}
\end{figure}

The Graph Convolutional Networks (GCNs \citep{kipf2016semi}) is a deep learning framework tailored for the processing of graph-structured data. It is a semi-supervised learning method based on graph structure that aggregates and propagates node features in the graph structure, thus deriving node representations. The Multi-GraphConv (M-G) Layer is a densely connected graph structure data processing module that is constructed based on GCNs. Illustrated in Figure~\ref{fig:fig3}, the $n$-th sublayer's output within the Multi-GraphConv (M-G) Layer serves as the subsequent $l-n$ sublayers' input, with the $N$-th layer receiving an aggregation of all output features from the initial $n-1$ layers.

Initially, the linear transformation of the adjacency matrix $\tilde{\boldsymbol{A}}^{\left(\text {head }_{\mathrm{i}}\right)}$ on the input features is computed for each head $i$. Subsequently, the impact of neighboring nodes' features on the present node is determined for each layer $l$, and the current node's features are consolidated with the previous layer's output:

\begin{align}
\begin{aligned}
\scalebox{0.87}{$
\tilde{H}^{(i,l)}=ReLU\left((\operatorname{\tilde{\boldsymbol{A}}^{\left(\text {head }_{\mathrm{i}}\right)}x})^{(i)} W^{(i, l)}\right) + \tilde{H}^{(i, l-1)}
$}
\end{aligned}
\end{align}where, $L$ is the quantity of layers in the graph convolution layer, $W^{(i,l)}$ is the weight parameter of the $i$-th head in the $l$-th layer.

The feature representation resulting from the output of each head is combined to form the final output of this layer:

\begin{align}
\begin{aligned}
\scalebox{0.87}{$
\tilde{H}^{(\mathbf{{MultiHead}})}={Concat}\left(\tilde{H}^{(1, l)}, \ldots, \tilde{H}^{(i, l)}\right) W^O
$}
\end{aligned}
\end{align}where $W^O \in \mathbb{R}^{hd \times d_{\text{model}}}$ is the weight parameter of the linear transformation applied to the ultimate output.

\subsubsection{Transformer-enc Layer}
The Transformer-enc layer is composed of multiple encoder layers stacked together. These encoder layers bear resemblance to the encoder layers delineated in the transformer model introduced by \citet{vaswani2017attention}. However, distinctively, our approach involves solely utilizing the output generated by the final three layers of the Encoder for the purpose of averaging. Each encoder layer incorporates self-attention mechanism and Feedforward Neural Network (FFN). This module facilitates the derivation of hidden representations of entities and an attention distribution matrix that are used as input for subsequent layers. The calculations can be outlined as follows:
\begin{align}
\begin{aligned}
\scalebox{0.87}{$\operatorname{self-Att}(\tilde{H}_Q, \tilde{H}_K, \tilde{H}_V)={softmax}\left(\frac{\tilde{H}_Q \tilde{H}_K^T}{\sqrt{d_k}}\right) \tilde{H}_V$}
\end{aligned}
\end{align}
\vspace{-5.5ex}
\begin{align}
\begin{aligned}
\scalebox{0.87}{$
La=\operatorname{LayerNorm}(\tilde{H}+\operatorname{self-Att}(\tilde{H}))$}
\end{aligned}
\end{align}
\vspace{-5.5ex}
\begin{align}
\begin{aligned}
\scalebox{0.87}{$
(\tilde{H},\tilde{A})=La+\operatorname{FFN}(La)$}
\end{aligned}
\end{align}
Where $\tilde{H}_Q, \tilde{H}_K, \tilde{H}_V$ is the query, key, and value representations obtained from the linear transformation of $\tilde{H}$, $d_k$ is the dimension of the attention head. $\text{LayerNorm}$ is the layer normalization operation.

\subsubsection{Collaborative Prediction Layer}

The local context extraction methodology, as described by \citet{zhou2021document}, is employed to ascertain the importance of individual tokens in relation to the entity pair $( Es,Eo)$, which is interpreted as the sentence-level importance. Erecting on this base, document-level importance was deduced by apportioning diverse attention weights in accordance to the contribution of each sentence within the document to the prediction of entity relations, and by establishing a fixed threshold. Sentences that exceed this threshold are selected as evidence sentences. The sentence-level importance $\boldsymbol{q_i}^{(Es, Eo)}$ and document-level importance $\boldsymbol{p_j}^{(Es, Eo)}$ can be computed as follows:
\begin{align}
\begin{aligned} 
\scalebox{0.95}{$
\boldsymbol{q_i}^{(Es, Eo)} =\sum_{i=1}^{\tilde{H}} \boldsymbol{\tilde{A}}_{Es} \cdot \boldsymbol{\tilde{A}}_{Eo}$}
\end{aligned}
\end{align}
\vspace{-4ex}
\begin{align}
\begin{aligned}
\scalebox{0.95}{$
\boldsymbol{p_j}^{(Es, Eo)}=\sum_{j=1}^{l} \boldsymbol{q_i}^{(Es, Eo)}$}
\end{aligned}
\end{align}

\begin{figure}[t]
\centering 
  \includegraphics[width=\columnwidth]{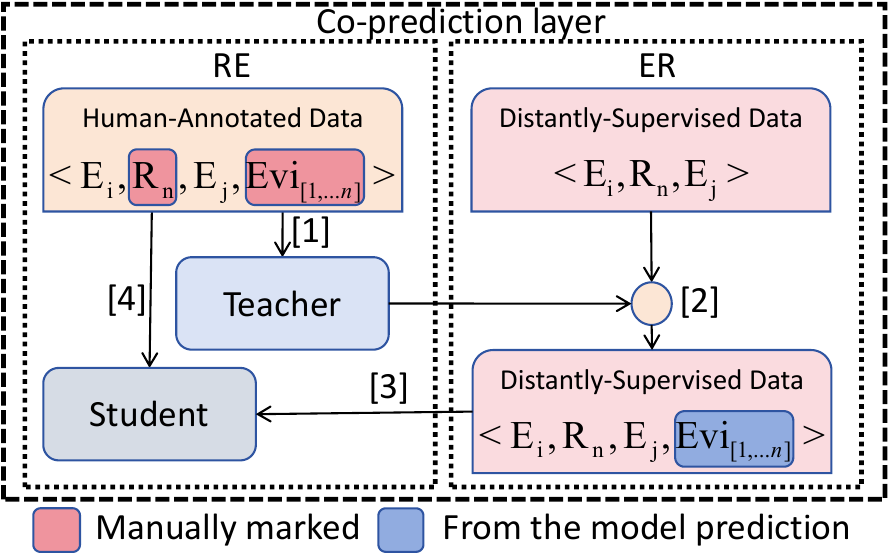}
  \caption{Step diagram of Co-prediction for RE and ER.}
  \label{fig:fig4}
\end{figure}

We apportion more attention to evidence sentences and less to non-evidence sentences through evidence supervision to further coordinate the prediction results of document-level relation extraction. As depicted in Figure~\ref{fig:fig4}, we train a teacher model on the Human-Annotated Data (which encompasses relation labels and evidence sentences) of DocRED (Step 1). We utilize the trained teacher model to predict the entity relations and evidence sentence distribution in Distantly-Supervised Data (which encompasses relation labels but lacks evidence sentences) (Step 2). Subsequently, we train a student model on the Distantly-Supervised Data that incorporates evidence sentences (Step 3), and retrain the student model using Human-Annotated Data (Step 4). 
Additionally, we define a row vector ${z}^{(Es, Eo)}$ consisting only of 0s and 1s, generated based on Human-Annotated Data. This vector indicates whether each sentence is an evidence sentence for the relation triples: $1$ if it is, and $0$ if it is not.
\begin{align}
\begin{aligned}
\scalebox{0.90}{$
\boldsymbol{z}^{(Es, Eo)}=\sum_{sent=1}^{L} \boldsymbol{z}^{(Es, Eo)} / \mathbf{1}^{\boldsymbol{\top}} \sum_{sent=1}^{L} \boldsymbol{z}^{(Es, Eo)}
$}
\end{aligned}
\end{align}
where $\mathbf{1}$ is the row vector composed of all $1$, $L$ is the total count of sentences contained within the document. $\boldsymbol{z}^{(Es, Eo)} $ and $\boldsymbol{p_j}^{(Es, Eo)}$ are mainly used in conjunction with Kullback Leibler (KL) divergence for loss calculation.
\vspace{-1ex}
\subsection{Classification Layer}

We begin with the computation of a weighted average of entity importance at the sentence-level $\boldsymbol{q_i}^{(Es, Eo)}$, and subsequently cascading it with the previous entity representation. Following this, we apply the $tanh$ activation function to normalize the input to range between $(-1, +1)$, resulting in the contextual representation of the two associated entities. The computation is elaborated as:
\vspace{-1ex}
\begin{align}
\begin{aligned} & 
\scalebox{0.80}{$
\boldsymbol{c}^{Es}=\tanh \left(\boldsymbol{W}^{Es}\left[\boldsymbol{e_{\text {emb}}}^{Es} ; \boldsymbol{\tilde{H}^{\boldsymbol{\top}}}\boldsymbol{q_i}^{(Es, Eo)} \right]+\boldsymbol{b}^{Es}\right) $}
\\ & 
\scalebox{0.80}{$\boldsymbol{c}^{Eo}=\tanh \left(\boldsymbol{W}^{Eo}\left[\boldsymbol{e_{\text {emb}}}^{Eo} ; \boldsymbol{\tilde{H}^{\boldsymbol{\top}}}\boldsymbol{q_i}^{(Es, Eo)}\right]+\boldsymbol{b}^{Eo}\right)$}
\end{aligned}
\end{align}where $\boldsymbol{W}^{Es}, \boldsymbol{W}^{Eo} \in \mathbb{R}^{d \times 2 d} $ and $\boldsymbol{b}^{Es}, \boldsymbol{b}^{Eo} \in \mathbb{R}^d$.

Finally, apply the grouped bilinear classifier proposed by \citet{zheng2019learning} to calculate the relation category scores.
$\boldsymbol{Score}^{(Es, Eo)}=\boldsymbol{c}_{Es}^{\boldsymbol{\top}} \mathrm{W}_{Rn} \boldsymbol{c}_{Eo}+\boldsymbol{b}_{Rn}$,
The possibility of the entity pair $(Es, Eo)$ possessing a relation Rn is computed thusly:
\scalebox{0.8}{$\boldsymbol{\mathrm{P}\left(Rn \mid Es, Eo\right)}\boldsymbol={Sigmoid}\left(\boldsymbol{Score}^{(Es, Eo)}\right)$}.

\section{Experiments}

\begin{table*}[t]
\centering
\resizebox{0.8\textwidth}{!}{
    \begin{tabular}{llcccclccc}
    \Xhline{1.5pt}
    && \multicolumn{3}{c}{ Dev } & & \multicolumn{4}{c}{ Test } \\
    \cline { 3 - 5 } \cline { 7 - 9 } 
    Category&\multirow{1}{*}{Model (With BERT$_{\text {base}}$)} & Ign-$F 1$ & $F 1$ & Evi-$F 1$ & & Ign-$F 1$ & $F 1$ & Evi-$F 1$  \\
    \hline
    \hline
    &$\bullet$ \multirow{1}{*}{\textbf{without Distant Supervision}}\\
    \hdashline
    
    \multirow{2}{*}{Sequence}
    &CNN \citep{yao2019docred} & 41.58 & 43.45 & -  & & 40.33 & 42.26 & - \\
    &BiLSTM \citep{yao2019docred} & 48.87 & 50.94 & -  & & 48.78 & 51.06 & -\\
    \hdashline
    \multirow{5}{*}{Graph}
    
    &GAIN \citep{zeng2020double}& 59.14 & 61.22& - & & 59.00 & 61.24 & - \\
    &MRN \citep{li2021mrn}& 59.74 & 61.61& - & & 59.52 & 61.74 & - \\
    &DocuNet \citep{zhang2021document} & 59.86 & 61.83 & -  & & 59.93 & 61.86 & -\\
    &GTN \citep{zhang2023exploring} & 60.86 & 62.73 & -  & & 60.77 & 62.75 & -\\
    &SD-DocRE \citep{zhang2023exploringa} & 60.85 & 62.81 & -  & & 60.91 & 62.85 & -\\
    \hdashline
    \multirow{4}{*}{Tranformer}
    &ATLOP \citep{zhou2021document} & 59.22 & 61.09 & -  & & 59.31 & 61.30 & -\\
    &EIDER \citep{xie2022eider} & 60.51 & 62.48 & 50.71 & & 60.42 & 62.47 & 51.27\\
    &SAIS \citep{xiao2022sais} & 59.98 & 62.96 & 53.70  & & 60.96 & 62.77 & 52.88\\
    &DREEAM \citep{ma2023dreeam} & 60.51 & 62.55 & 52.08  & & 60.03 & 62.49 & 51.71\\
    \hdashline
    \multirow{2}{*}{Teacher}
    &GEGA-single (Ours) & 59.98$_{\pm0.12}$ & 61.95$_{\pm0.12}$ & 52.19$_{\pm0.15}$  & & 59.31 & 61.52 & 51.90\\
    &GEGA-fusion (Ours) & 60.55$_{\pm0.08}$ & 62.65$_{\pm0.08}$ & -  & & 60.11 & 62.53 & -\\
    
    \hline
    \hline
    
    &$\bullet$ \multirow{1}{*}{\textbf{with Distant Supervision}}\\
    \hdashline
    \multirow{1}{*}{Graph}
    &AA \citep{lu2023anaphor} & 61.31 & 63.38 & -  & & 60.84 & 63.10 & -\\
    \hdashline
    \multirow{2}{*}{Tranformer}
    &KD-DocRE \citep{tan2022document} & 62.62 & 64.81 & - & & 62.56 & 64.76 & -\\
    
    &DREEAM \citep{ma2023dreeam} & 63.92 & 65.83 & 55.68  & & 63.73 & 65.87 & 55.43\\
    \hdashline
    \multirow{2}{*}{Student}

    &GEGA-single (Ours) & 64.02$_{\pm0.15}$ & 65.83$_{\pm0.15}$ & $\mathbf{56.09}$$_{\pm0.18}$  & & 63.82 & 65.85 & $\mathbf{55.89}$\\
    
    &GEGA-fusion (Ours) & $\mathbf{64.26}$$_{\pm0.13}$ & $\mathbf{66.38}$$_{\pm0.13}$ & -  & & $\mathbf{63.90}$ & $\mathbf{66.31}$ & -\\

    \Xhline{1.5pt}
    \end{tabular}%
    }
    \caption{Experimental results (\%) for the dev and test set of DocRED. Using BERT-base as a pre-trained language model. The best score has been displayed in bold. The scores of other models refer to their respective papers.
      }
    \label{tab:DocRED}
\end{table*}

\subsection{Dataset and Evaluation}
DocRED\footnote{\url{https://github.com/thunlp/DocRED}} \citep{yao2019docred} is a benchmark dataset for document-level relation extraction tasks, released by Tsinghua University. DocRED comprises numerous documents from Wikipedia and Wikidata, each annotated with entities, relations between entities, and evidence sentences that support relation triples. It serves as the predominant benchmark for DocRE model training and evaluation.

Re-DocRED\footnote{\url{https://github.com/tonytan48/Re-DocRED}} \citep{tan2022revisiting} and Revisit-DocRED\footnote{\url{https://github.com/AndrewZhe/Revisit-DocRED}} \citep{huang2022does} are modified datasets of DocRED. They supplement a large number of relation triples to solve the problems of incomplete annotations, coreferential errors, and inconsistent logic in docred. Annotation quality has high accuracy and consistency, which provides a more reliable benchmark for DocRE-model training and evaluation.

We assess GEGA using an Nvidia Tesla V100 16GB GPU and evaluate it with F1, Ign-F1, and Evi-F1 metrics. Ign-F1 represents the calculated F1 score attained by excluding relational facts present in both the training and development/testing datasets. 
Evi-F1 serves as a significant measure for assessing the performance of ER and constitutes a new benchmark for assessing the quality of relation extraction models.
\subsection{Single and Fusion}
In the task of RE, the most ideal scenario is that the evidence sentence set of the dataset already contains all contextual information necessary to predict entity relations, thereby enabling accurate relation prediction results based solely on the evidence sentence set. However, manually labeled data and distant supervision data often fall short in this regard. Therefore, it is necessary to extract contextual information from the entire document to predict entity relations. We divide the above problem into two evaluation methods: (1) Single: extract entity relations from the entire document and obtain the corresponding prediction scores; (2) Fusion: predict entity relations based on a collection of evidence sentences and combine the prediction results with those from the Single method. This is similar to the Fusion of Evidence approach in \citet{xie2022eider}.

\subsection{Compared Methods}
To ensure a fair comparison of the performance of DocRE baselines, we compare our model with three state-of-the-art methods, all using BERT-base as the pre-trained language model (PLM), which are:
(1) Sequence-based methods:
CNN \citep{yao2019docred},
LSTM \citep{yao2019docred} ,
BiLSTM \citep{yao2019docred}. 
(2) Graph-based methods:
GAIN \citep{zeng2020double},
MRN \citep{li2021mrn},
DocuNet \citep{zhang2021document},
GTN \citep{zhang2023exploring},
SD-DocRE \citep{zhang2023exploringa},
AA \citep{lu2023anaphor}. 
(3) Transformer-based methods:
ATLOP \citep{zhou2021document},
EIDER \citep{xie2022eider},
SAIS \citep{xiao2022sais},
PRiSM \citep{choi2023prism},
DREEAM \citep{ma2023dreeam}.
On this foundation, we categorize the above methods into two major classes: without Distant Supervision and with Distant Supervision.

\begin{table*}
    \centering
    \resizebox{0.75\textwidth}{!}{
    \begin{tabular}{llccccccc}
    \Xhline{1.5pt}
    && \multicolumn{2}{c}{ Dev } & & \multicolumn{2}{c}{ Test } \\
    \cline { 3 - 4 } \cline { 6 - 7 } 
    Category&\multirow{1}{*}{Model (With BERT$_{\text {base}}$)} & Ign-$F 1$ & $F 1$ && Ign-$F 1$ & $F 1$ \\
    \hline
    \hline
    
    \multirow{3}{*}{Graph}
    &GAIN-BERT \citep{zeng2020double} & 71.99 & 73.49  & & 71.88 & 73.44 \\
    &DocuNet-BERT \citep{zhang2021document} & 73.68 & 74.65  & & 73.60 & 74.49 \\
    &GTN-BERT  \citep{zhang2023exploring} & 75.03 & 75.85  & & 74.85 & 75.77 \\
    \hdashline
    \multirow{4}{*}{Tranformer}
    &ATLOP-BERT \citep{zhou2021document} & 73.35 & 74.22  & & 73.22 & 74.02 \\
    &KMGRE-BERT  \citep{jiang2022key} & 73.33 & 74.44  & & 73.39 & 74.46 \\
    &KD-DocRE-BERT \citep{tan2022document} & 73.76 & 74.69  & & 73.67 & 74.55 \\
    &PRiSM-BERT \citep{choi2023prism} & 72.92 & 74.25  & & 72.35 & 73.69 \\
    \hdashline
    
    \hdashline
    
    \hdashline
    
    \hdashline
    \multirow{2}{*}{Teacher}
    &GEGA-single (Ours) & 73.69$_{\pm0.06}$ & 74.53$_{\pm0.06}$& &73.42$_{\pm0.05}$ & 74.21$_{\pm0.03}$\\
    &GEGA-fusion (Ours) & 76.06$_{\pm0.07}$ & 77.41$_{\pm0.07}$& &75.28$_{\pm0.03}$ & 76.61$_{\pm0.03}$\\
    \hdashline
    \multirow{2}{*}{Student}
    &GEGA-single (Ours) & 75.72$_{\pm0.06}$ & 76.64$_{\pm0.06}$& &75.25$_{\pm0.05}$ & 76.21$_{\pm0.05}$\\
    &GEGA-fusion (Ours) & $\mathbf{77.11}$$_{\pm0.08}$ & $\mathbf{78.37}$$_{\pm0.08}$& &$\mathbf{76.43}$$_{\pm0.06}$ & $\mathbf{77.74}$$_{\pm0.07}$\\
    
    \Xhline{1.5pt}
    \end{tabular}%
    }
    \caption{\label{tab:Re-DocRED}
    Performance (\%) on the dev/test set of Re-DocRED. We use the same presentation method as Table~\ref{tab:DocRED}. Other model results are replicated from the academic paper \citep {zhang2023exploring}.
    }
\end{table*}

\section{Results and Analyses}

We test the trained student and teacher models in both the Single and Fusion stages, and report the results on DocRED, Re-DocRED and Revisit-DocRED, where the results on Revisit-DocRED are moved to appendix.
\subsection{Results on DocRED}
Table~\ref{tab:DocRED} indicates that GEGA achieves superior Ign-F1 and F1 metrics compared to the established DocRE Baselines on both the development set and the test set. The single stage of the student model has achieved performance levels comparable to the leading DREEAM \citep{ma2023dreeam}. Notably, the fusion stage of the student model achieved the highest recorded scores, surpassing DREEAM by 0.34\% (Ign F1) and 0.55\% (F1) on the development set, as well as by 0.17\% (Ign F1) and 0.44\% (F1) on the test set.

GEGA also performs well in the test of the new benchmark Evi-F1, surpassing the previous most advanced DREEAM by 0.41\% (Evi-F1) and 0.46\% (Evi-F1) on the development set and test set respectively. In the table, it is evident that the Transformer-based and Graph-based models outperform the Sequence-based ones, validating the rationality behind integrating infographics with the Transformer.

\subsection{Results on Re-DocRED}
Table~\ref{tab:Re-DocRED} presents the feedback outcomes of GEGA on the development and test sets of RE-DocRED, demonstrating that our GEGA achieves state-of-the-art results compared to other methods utilizing BERT-base as a pre-trained language model. Notably, GEGA has outperformed all other methods in the table during the fusion stage without Distant Supervision (Teacher). GEGA secured the highest Ign F1 and F1 scores in the fusion stage with Distant Supervision (Student), with improvements of 2.08\% (Ign F1) and 2.52\% (F1) respectively on the development set over the second-place GTN-BERT \citep{zhang2023exploring}, and by 1.58\% (Ign F1) and 1.97\% (F1) respectively on the test set.

\subsection{Effect Analysis of GCNs and ER}
Based on the test scores on DocRED and Re-DocRED, we observe that graph-based models such as DocuNET \citep{zhang2021document} and GTN-BERT \citep{zhang2023exploring} have achieved superior performance in the field. Graphs have an advantage in conveying document-level contextual information, so we used grid search to select a 2-layer GNNs to guide multiple attention maps. Additionally, DREEAM \citep{ma2023dreeam} is the first method to enhance relation extraction performance purely through evidence-guided attention, it has already achieved excellent scores on the DocRED set. By integrating GCNs with evidence retrieval, we further improved its scores by 0.41\% (Evi-F1) and 0.46\% (Evi-F1) on the dev and test sets, respectively. Therefore, we conclude that GCNs and ER significantly enhance performance in the relation extraction task.

\subsection{Ablation Studies}

\begin{table}
\resizebox{\columnwidth}{!}{
\setlength{\tabcolsep}{0pt}
    \begin{tabular}{lcccc}
    \Xhline{1.5pt}
    && \multicolumn{1}{c}{ DocRED-Dev } & \\
    \cline { 2 - 4 }
    \multirow{1}{*}{Model (With BERT$_{\text {base}}$)} &  Ign-$F 1$ & $F 1$ & Evi-$F 1$ \\
    \hline
    \hline
    
    $\bullet$ \multirow{1}{*}{\textbf{GNNs layer ablation}}\\
    GEGA-single & $\mathbf{64.02}$ & $\mathbf{65.83}$ & $\mathbf{56.09}$\\

    — Attention Concentration Layer & 63.37&65.34& 55.69 \\
    — Multi-GraphConv Layer & 62.94 & 64.31&55.19 \\
    — Transformer-enc Layer & 63.87 & 65.66&55.93 \\
    \hdashline
    
    \hdashline
    
    \hdashline
    
    \hdashline
    $\bullet$ \multirow{1}{*}{\textbf{Training phase ablation}}\\
    GEGA-single & $\mathbf{64.02}$ & $\mathbf{65.83}$ & $\mathbf{56.09}$\\

    — self-training & 62.19 & 63.45   & 53.98 \\
    — fine-tuning & 63.91 & 65.80   & 55.63 \\
    — Distant Supervision-training & 59.98 & 61.95 & 52.19 \\
    
    \Xhline{1.5pt}
    \end{tabular}%
    }
    \caption{\label{tab:ablation}
    Ablation analysis on DocRED-Dev.
  }
\end{table}

We conducted ablation experiments were conducted on the development set to analyze the GNNs layer and Training phase of GEGA. The single phase of GEGA (Student) was utilized as the test benchmark. The results of the score post-ablation of each part are presented in Table~\ref{tab:ablation}. Initially, we removed the Attention Concentration Layer, which resulted in a minor decline in performance. Subsequently, upon removing the Multi-GraphConv Layer, a significant performance deterioration was observed, implying the importance of constructing multiple attention distribution graphs for relation extraction. Upon removal of the Transformer-enc Layer, we noted a relatively minor decline in performance. We speculate that this may be related to using Transformer based BERT as PLM.

Additionally, during the training process, we performed ablation on the self-training, fine-tuning, and Distant Supervision-training stages, in order to further analyze their impact. The results indicate that when the ER self-training phase is omitted, performance declines, whereas the absence of the fine-tuning stage did not lead to a noticeable decline in performance. Further more, omitting the Distant Supervision-training stage caused severe performance degradation. These findings highlight the effectiveness of our ER method in enhancing relation extraction.


  \vspace{-1ex}
\section{Conclusion}
We propose GEGA, the first model to employ GCNs and ER jointly guided attention to enhance DocRE. We validate the superiority of our model on three widely used datasets: DocRED, Re-DocRED and Revisit-DocRED. GEGA is trained using parallel computing in both fully supervised and semi-supervised settings, without incurring additional overhead, making it convenient for use in the era of Large Language Models (LLMs). In the future, we aim to leverage the scalability of GEGA and apply it to a broader range of scenarios, including entity recognition, event extraction, and more.

\section{Limitations}
The model GEGA is subject to two limitations. Firstly, when utilizing Multi-GraphConv Layers to induce multiple fully connected attention distribution matrices, there is a possibility of generating one matrix that differs significantly from others in terms of weight distribution. This could lead to significant deviations in prediction results. We hypothesize that guiding the construction of multiple fully connected attention matrices using evidence information may reduce the occurrence of such undesirable situations, a conjecture that will be verified in future work.
Secondly, it is acknowledged that the relations between most entity pairs can be predicted based on the local context of the entities. However, our model utilizes evidence sentences retrieved from the entire document corpus, which are strongly correlated with the entity pairs of interest, rather than evidence sentences obtained specifically for individual relation triples. This approach may result in the model carrying more global contextual information while reducing the utilization of local context information.

\section{Ethics Statement}
Our proposed GEGA demonstrates outstanding scalability and applicability, serving as an excellent solution for both DocRE and DocER tasks. This method is evaluated solely on publicly available datasets, ensuring no compromise on individual privacy. Furthermore, we provide the source code implementation of GEGA to enable researchers to reproduce its performance authentically, fostering academic exchange in the field of DocRE.


\bibliography{custom}

\appendix
\section{Loss}
\begin{figure}
\centering 
  \includegraphics[width=\columnwidth]{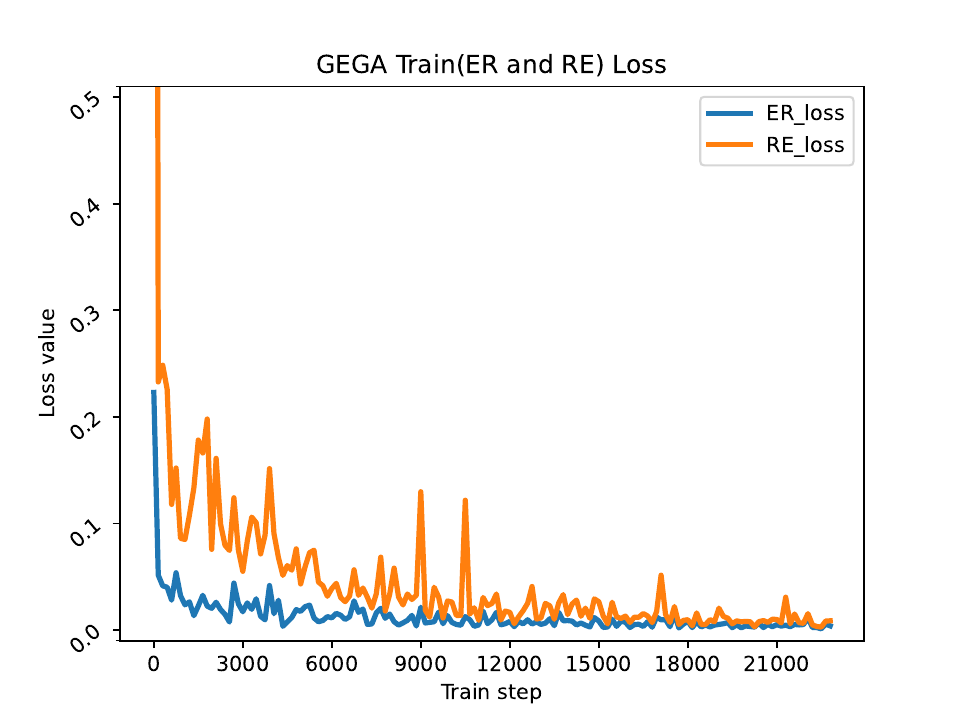}
  \caption{Loss value variation of the GEGA model trained on the DocRED dataset}
  \label{fig:loss}
\end{figure}

To accommodate the requirements of the RE and ER tasks within both the teacher model and the student model, distinct forms of loss function computation have been devised for the relation classification approach.The loss variation is shown in Figure~\ref{fig:loss}.

\subsection{RE loss:}
For the RE tasks of the two models previously described, we implement the Adaptive Thresholding Loss (ATL) as proposed by ATLOP. During the training phase, we use a threshold class (TH) to learn a threshold such that the logits of the positive class $\mathcal{RP}$ exceed it, and the logits of the negative class $\mathcal{RN}$ fall below it.

\begin{align}
\begin{aligned}
\mathcal{L}_{\mathrm{RE}}= &
\scalebox{0.95}{$
-\sum_{{Rn} \in \mathcal{RP}} \frac{\exp \left(Score_{Rn}^{(Es, Eo)}\right)}{\sum_{{Rn}^{\prime} \in \mathcal{RP} \cup\{\mathrm{TH}\}} \exp \left(Score_{{Rn}^{\prime}}^{(Es, Eo)}\right)} 
$}
\\ & 
\scalebox{0.95}{$-\frac{\exp \left(Score_{\mathrm{TH}}^{(Es, Eo)}\right)}{\sum_{{Rn}^{\prime} \in \mathcal{RN} \cup\{\mathrm{TH}\}} \exp \left(Score_{{Rn}^{\prime}}^{(Es, Eo)}\right)}$} &
\end{aligned}
\end{align}


\subsection{ER loss:}
The tasks targeted by the teacher model and the student model differ in detail. The teacher model is trained on Human-Annotated Data, which includes reliable manually annotated evidence sentences, while the student model is trained on Distantly-Supervised Data that incorporates evidence sentences identified through the teacher model's ER. Considering these distinctions, there is a requirement for specialized loss computation methods. Consequently, we propose both document-level and sentence-level loss calculations.

\textbf{Document-level Loss:}
By integrating the document-level importance distribution $\boldsymbol{p_j}^{(Es, Eo)}$ with the original manually annotated evidence sentences $\boldsymbol{z}^{(Es, Eo)}$ from the dataset, we induce a localized context representation that contributes significantly to RE. We use the Kullback-Leibler Divergence (KL divergence), a method for measuring the difference between two probability distributions within the same event space. In this paper, it is used to assess the degree of divergence between $\boldsymbol{p_j}^{(Es, Eo)}$ and $\boldsymbol{z}^{(Es, Eo)}$:
\begin{align}
\begin{aligned}
\mathcal{L}_{\mathrm{ER}}^{\text {doc}}=
\mathrm{KL}( \boldsymbol{z}^{(Es, Eo)} \| \boldsymbol{p_j}^{(Es, Eo)} )&\\
=\sum_{sent=1}^{L} \boldsymbol{z}^{(Es, Eo)} \log \frac{\boldsymbol{z}^{(Es, Eo)}}{\boldsymbol{p_j}^{(Es, Eo)}}
\end{aligned}
\end{align}

\textbf{Sentence-level Loss:}
We use the teacher model trained on Human-Annotated Data to perform ER testing on Distantly-Supervised Data, predicting its sentence-level evidence distribution $\boldsymbol{\tilde{q}}_{\boldsymbol{i}}^{(E s, E o)}$.
Thereafter, utilize Kullback-Leibler (KL) divergence to compute the difference in sentence-level probability distributions between the teacher model and the student model.
\begin{align}
\begin{aligned} & \mathcal{L}_{\mathrm{ER}}^{\mathrm{sent}}=\mathrm{KL}\left(\boldsymbol{q}_{\boldsymbol{i}}^{(E s, E o)} \| \boldsymbol{\tilde{q}}_{\boldsymbol{i}}^{(E s, E o)}\right) \\ & =\sum_{t=1}^l \boldsymbol{\boldsymbol{q}_{\boldsymbol{i}}}^{(E s, E o)} \log \frac{\boldsymbol{\boldsymbol{q}_{\boldsymbol{i}}}^{(E s, E o)}}{\boldsymbol{\tilde{q}}_{\boldsymbol{i}}^{(E s, E o)}}
\end{aligned}
\end{align}
Finally, we apply the prevalent weighted summation technique for document-level and sentence-level losses to equilibrate the losses of RE and ER, where $\lambda$ serving as a hyperparameter.
\begin{align}
\begin{aligned}\mathcal{L}=(1-\lambda)\mathcal{L}_{\mathrm{RE}}+\lambda \mathcal{L}_{\mathrm{ER}}
\end{aligned}
\end{align}

\section{Hyperparameter}
We ran tests on the above three datasets using different random seeds five times and reported the average test accuracy.
In the manuscript, a comprehensive Table~\ref{Hyperparameter} delineating the hyperparameters and configuration has been furnished, encapsulating pivotal settings utilized across the experimental trials. These hyperparameters, meticulously curated and fine-tuned, exert control over multifarious facets of the model's dynamics throughout the training and evaluation phases. 
\begin{table*}
    \centering
    \resizebox{0.78\textwidth}{!}{
    \begin{tabular}{lccccc}
    \Xhline{1.5pt}
    \multirow{1}{*}{Hyperparameter/Configuration} & teacher & student  & self-train  & finetune & evaluation\\
    \hline
    \hline

    train-file & annotated & - & distant & annotated&-\\
    test-file & - & distant & - & -&test\\
    dev-file & dev & - & dev & dev&-\\
    num-class & 97 & 97 & 97 & 97&97\\
    gradient-accumulation-steps & 1 & - & 2 & 1&-\\
    train-batch-size & 4 & - & 4 & 4&-\\
    test-batch-size & 8 & 4 & 8 & 8&8\\
    num-labels & 4 & 4 & 4 & 4&4\\
    evi-$\lambda$ & 0.1 & 0.1 & 0.1 & 0.1&-\\
    lr-transformer & 5e-5 & - & 3e-5 & 1e-6&-\\
    lr-added & - & - & - & 3e-6&-\\
    max-grad-norm & 1.0 & - & 5.0 & 2.0&-\\
    evi-thresh & 0.2 & 0.2 & 0.2 & 0.2&0.2\\
    warmup-ratio & 0.06 & - & 0.06 & 0.06&-\\
    num-train-epochs & 30.0 & - & 2.0 & 10.0&-\\
    eval-mode & - & - & - & - & single/fusion \\
    \Xhline{1.5pt}
    \end{tabular}%
    }
    \caption{\label{Hyperparameter}
        Hyperparameter/Configuration Settings for Training and Evaluation of GEGA.
  }
\end{table*}

\section{Results on Revisit-DocRED}
Table~\ref{tab:Revisit-DocRED} shows the test results of GEGA (Student) on Revisit-DocRED. According to the test method of \citet{huang2022does}, we trained on the DocRED training set, and then tested on the test set provided by Revisit-DocRED. It can be seen that GEGA has a great performance improvement compared with other methods.

\begin{table}
    \centering
    \resizebox{\columnwidth}{!}{
    \begin{tabular}{llccccccc}
    \Xhline{1.5pt}
     & \multicolumn{2}{c}{ Test } \\
    \cline { 2 - 3 }
    \multirow{1}{*}{Model (With BERT$_{\text {base}}$)} & Ign-$F 1$ & $F 1$\\
    \hline
    \hline
    
    CNN-BERT*  \citep{yao2019docred}& 29.70 & 30.04\\
    LSTM-BERT* \citep{yao2019docred} & 31.32 & 31.77 \\
    BiLSTM-BERT*  \citep{yao2019docred} & 32.50 & 32.91\\
    
    GAIN-BERT \citep{zeng2020double} & 41.27 & 41.64 \\
    ATLOP-BERT \citep{zhou2021document} & 41.62 & 41.90\\
    KMGRE-BERT  \citep{jiang2022key} & 42.78 & 43.16 \\
    KD-DocRE-BERT  \citep{tan2022document} & 43.22 & 43.68 \\
    GTN-BERT  \citep{zhang2023exploring} & 44.84 & 45.33 \\
    DREEAM-BERT*  \citep{ma2023dreeam} & 55.32 & 56.48 \\
    \hdashline
    
    \hdashline
    
    \hdashline
    
    \hdashline
    GEGA-single (Ours) & 45.34$_{\pm0.12}$ & 45.58$_{\pm0.10}$\\
    GEGA-fusion (Ours) & $\mathbf{55.89}$$_{\pm0.07}$ & $\mathbf{56.97}$$_{\pm0.05}$\\

    \Xhline{1.5pt}
    \end{tabular}
    }
    \caption{\label{tab:Revisit-DocRED}
    Performance (\%) on the test set of Revisit-DocRED. Results marked with * are obtained by our code reproduction. Other model results are replicated from the academic paper \citep {zhang2023exploring}.
    }
\end{table}

\begin{figure*}[t]
\centering
    \subfloat[Head count comparison of Multi-GraphConv layer.]           {\includegraphics[width=\columnwidth]{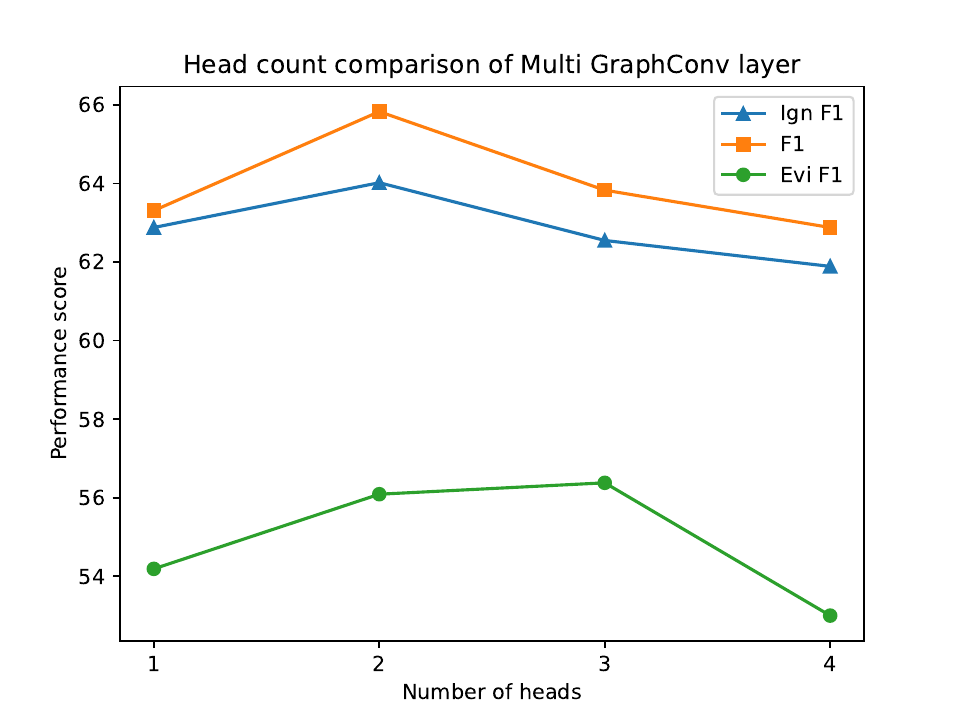}\label{fig:head}}
    \hfill
    \subfloat[Comparison of different layers of GNNs.]
    {\includegraphics[width=\columnwidth]{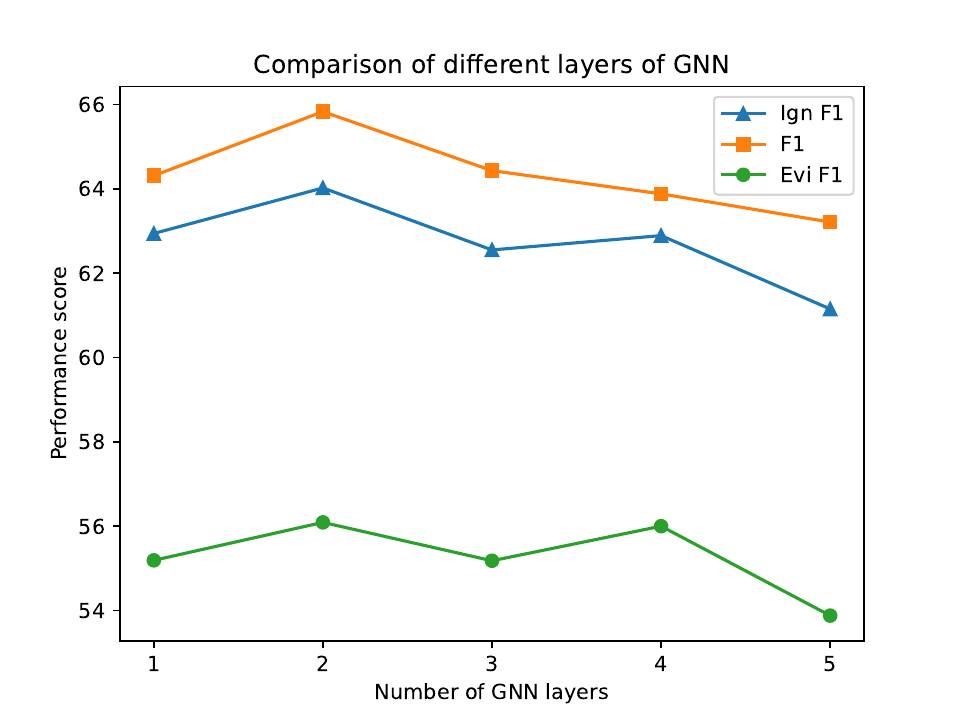}\label{fig:gnn}}
     \caption{Comparison of experimental results with different parameter settings of GEGA.}
    \label{fig:comparison}
\end{figure*}

\section{Supplementary Model Analysis}
Here, we conducted experimental analyses on the number of GNNs layers and the number of heads in the Multi-GraphConv Layer of GEGA. We selected the number of GNNs layers from $\{1, 2, 3, 4, 5\}$ and the number of heads from $\{1, 2, 3, 4\}$. The experimental results are shown in Figure~\ref{fig:comparison}. Ultimately, the analysis concluded that the performance is optimal when $layers$=$2$ and $heads$=$2$.



\section{Case Study}
\begin{figure*}[t]
\centering 
  \includegraphics[width=\textwidth]{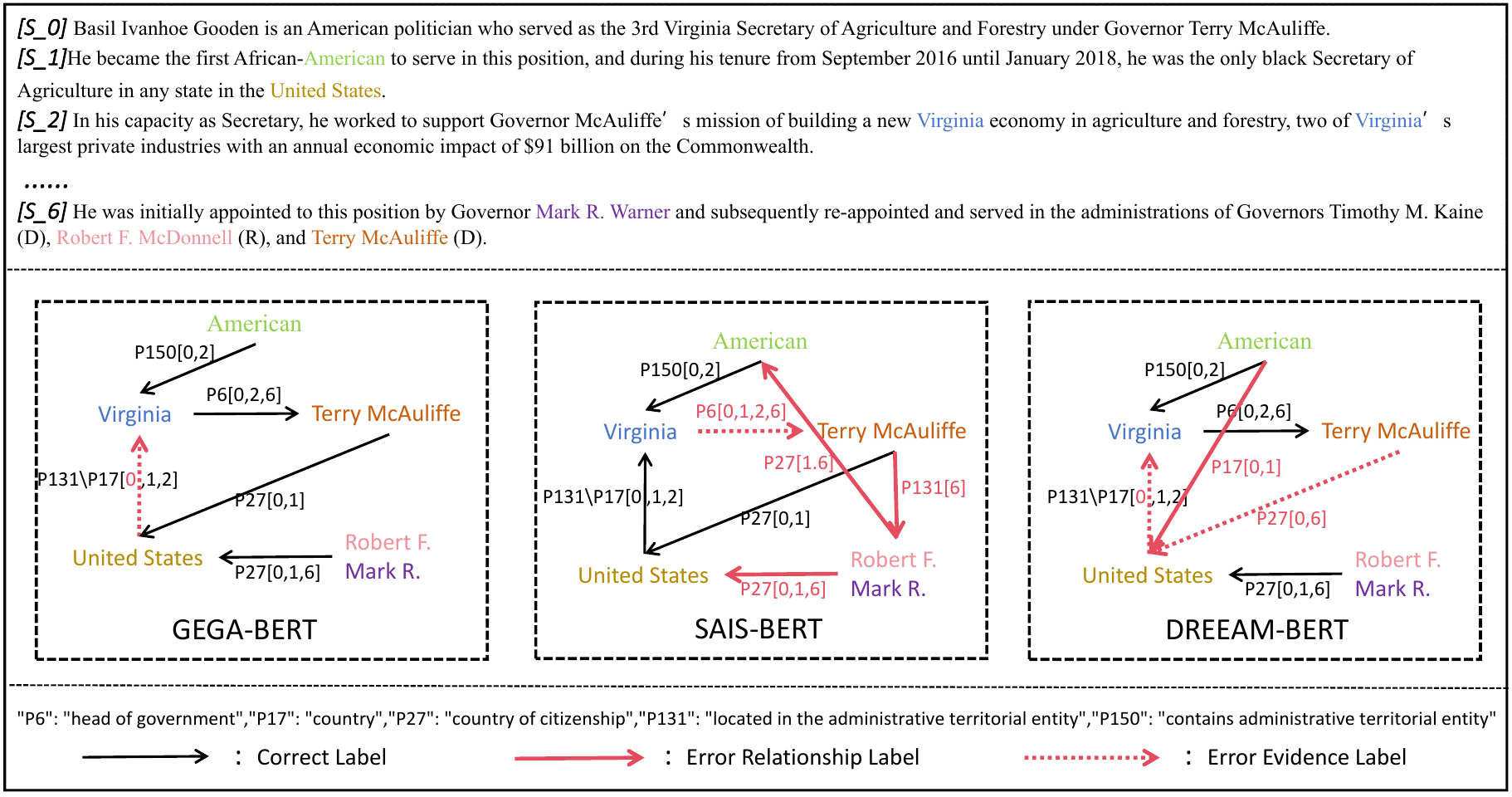}
  \caption{The comparison results of a case on three advanced models show that the entity is marked with special color, $P_{number}$ is the relation label, and the red arrow is the prediction error.}
  \label{fig:case}
\end{figure*}

To further demonstrate the superior performance of GEGA, we extracted a case from the DocRED dataset and used it to compare the performance of GEGA with two other advanced methods (SAIS and DREEAM).
\subsection{RE}
From Figure~\ref{fig:case}, we observe five types of relations among five entities. SAIS correctly identified four types of relations but overlooked the \textit{"country of citizenship"} relation between "$Robert\: F./Mark\: R.$" and "$United\:States$" Additionally, it incorrectly identified a \textit{"country of citizenship"} relation between "$Robert\: F./Mark\: R.$" and "$American$" as well as a \textit{"located in the administrative territorial entity"} relation with "$Terry\: McAuliffe$" While DREEAM correctly identified all the existing relations, it excessively identified a \textit{"country"} relation between "$American$" and "$United\: States$" In contrast, GEGA perfectly extracted the correct relational network in this case.
\subsection{ER}
During evidence retrieval, both GEGA and DREEAM labeled the evidence source for the relation between $United\: States$ and $Virginia$ as [S1, S2], missing [S0]. Additionally, DREEAM incorrectly labeled the evidence information for the relation between $Terry\: McAuliffe$ and $United\: States$ as [S0, S6]. SAIS incorrectly labeled the evidence information for the relation between $Virginia$ and $Terry\: McAuliffe$ as [S0, S1, S2, S6]. We believe that GEGA's superior performance is also attributed to its better ER performance.

\end{document}